\documentclass[letterpaper, 10 pt, conference]{ieeeconf}  

\IEEEoverridecommandlockouts                              

\usepackage{graphicx} 
\usepackage{mathptmx} 
\usepackage{amsmath} 
\usepackage{amssymb} 
\usepackage{cite}
\usepackage{multirow}
\usepackage{xcolor} 
\usepackage{siunitx}
\usepackage{algorithm}
\usepackage{algorithmic}
\usepackage{booktabs}
\usepackage{stfloats}
\usepackage{bm}
\usepackage{mathrsfs}
\usepackage{makecell}
\usepackage{times} 
\usepackage{mathrsfs}

\usepackage{subcaption} 

\DeclareMathAlphabet{\bbold}{U}{bbold}{m}{n}
\newcommand{\id}{\ensuremath{\bbold{1}}}
\usepackage{colortbl}
\definecolor{greenbetter}{RGB}{33, 217, 82}
\definecolor{greenbetter2}{RGB}{147, 219, 166} 
\definecolor{yellowneutral}{RGB}{255,255,128}
\definecolor{redworse}{RGB}{242,182,182}

\newtheorem{remark}{Remark}

\usepackage[pagebackref,breaklinks,colorlinks]{hyperref}

\usepackage{cleveref}

\title{\LARGE \bf $f$COP: Focal Length Estimation from Category-level Object Priors}

\author{Xinyue Zhang$^{1*}$, Jiaqi Yang$^{1*}$, Xiangting Meng$^{1}$, Abdelrahman Mohamed$^{2}$ and Laurent Kneip$^{1}$
\thanks{$^{1}$Xinyue Zhang, Jiaqi Yang, Xiangting Meng and Laurent Kneip are with the School of Information Science and Technology, Computer Science,
        ShanghaiTech University, Pudong Shanghai, China
        {\tt\small \{zhangxy11, yangjq, mengxt, lkneip\}@shanghaitech.edu.cn}}%
\thanks{$^{2}$Abdelrahman Mohamed are with Rembrand, Palo Alto, CA 94306, USA {\tt\small abdo@rembrand.com} }
\thanks{$^{*}$ have the equal contribution}
\thanks{This work has been submitted to the IEEE for possible publication. 
Copyright may be transferred without notice, after which this version may no longer be accessible.}
}

\begin{document}

\maketitle
\thispagestyle{empty}
\pagestyle{empty}
\begin{abstract} 
In the realm of computer vision, the perception and reconstruction of the 3D world through vision signals heavily rely on camera intrinsic parameters, which have long been a subject of intense research within the community.  In practical applications, without a strong scene geometry prior like the Manhattan World assumption or special artificial calibration patterns, monocular focal length estimation becomes a challenging task. In this paper, we propose a method for monocular focal length estimation using category-level object priors. Based on two well-studied existing tasks: monocular depth estimation and category-level object canonical representation learning, our focal solver takes depth priors and object shape priors from images containing objects and estimates the focal length from triplets of correspondences in closed form. Our experiments on simulated and real world data demonstrate that the proposed method outperforms the current state-of-the-art, offering a promising solution to the long-standing monocular focal length estimation problem.

\end{abstract}    
\section{Introduction}
\label{sec:intro}

Traditionally, the solution of 3D vision-based perception problems heavily relies on existing knowledge about the intrinsic parameters of a camera. For example, in structure from motion or visual SLAM frameworks, the camera calibration is often identified upfront using dedicated camera calibration procedures. In order to better understand the need for calibration parameters, let us recite the \textit{projective reconstruction theorem}~\cite{hartley2003multiple}: 

\vspace{0.5cm}
\textit{From several images of a scene and the coordinates of corresponding points identified in the different images, it is possible to construct a 3-dimensional point-cloud model of the scene, and compute the camera locations. From uncalibrated images the model can be reconstructed up to an unknown projective transformation, which can be upgraded to a Euclidean model by adding or computing calibration information.}
\vspace{0.5cm}

In the case of multiple images, the augmentation to the calibrated case can be facilitated if the assumptions of a perspective camera with principal point located at the center of the image is sufficiently valid. As demonstrated by Li et al.~\cite{li2006simple},
in this case the remaining unknown focal length parameter can be identified from simple multi-view geometry. Even under the same assumptions, a more challenging case however occurs when only a single image is given, which is the topic of the present paper.

In the single image scenario, the community has proposed several traditional methods that primarily rely on prior scene assumption. Examples are given by known patterns in the scene (i.e. calibration targets), objects or scenes of at least partially known geometry, or more high-level assumptions such as Manhattan world properties. While certainly interesting, these traditional approaches are sometimes not a viable choice given that the assumptions they rely on can be considered strong. The goal of the present paper is to leverage perhaps more generally applicable priors that are nowadays given by learning-based methods. As we will show, it is possible to utilize monocular depth prediction and category-level object priors to relatively simply extract knowledge about the focal length.

\begin{figure}[t!]
    \centering
    \includegraphics[width=0.99\linewidth]{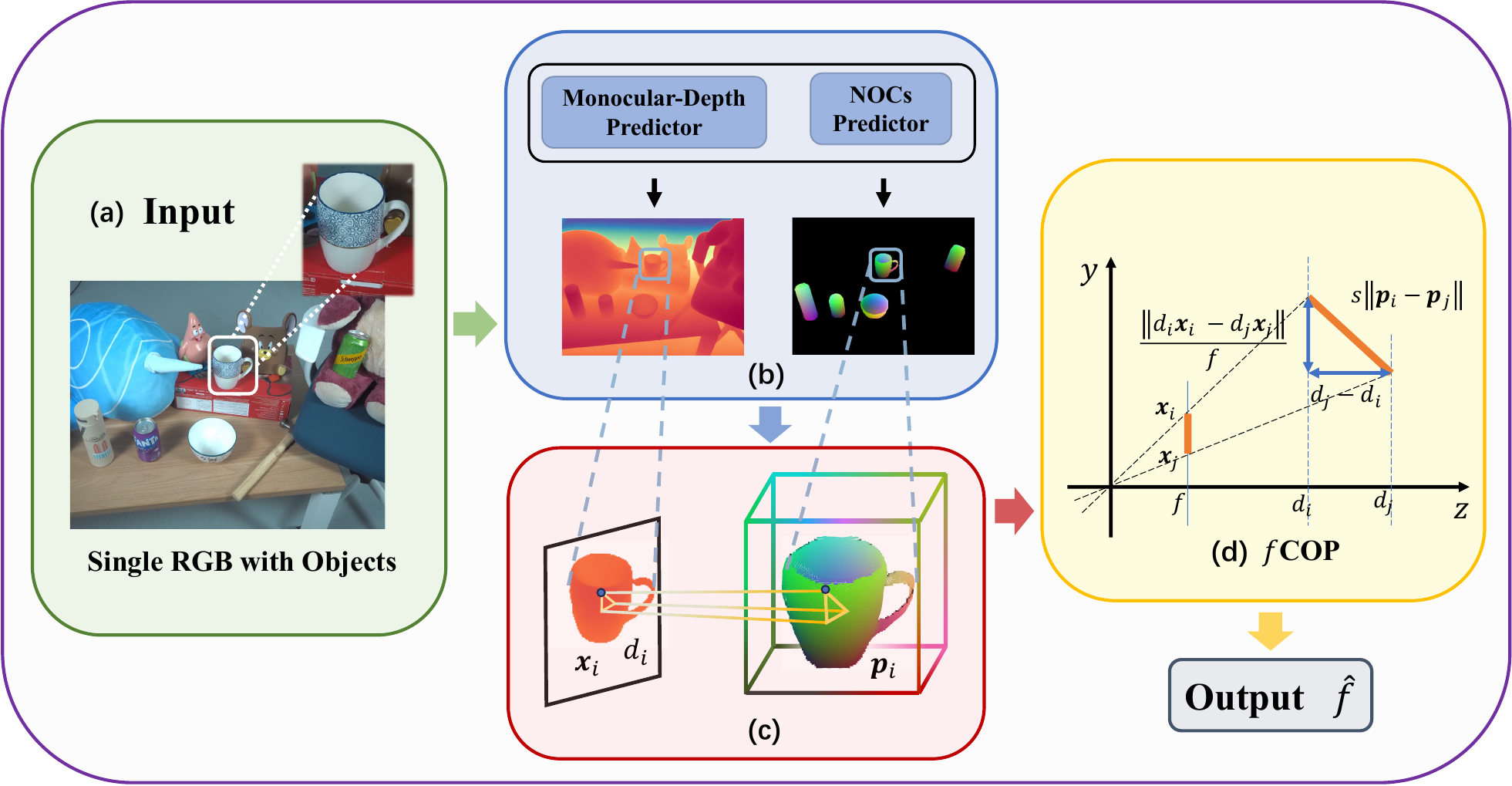}
    \caption{Overview of the pipeline for focal length estimation using category-level object priors. 
    (a) The input to our approach is an RGB image containing objects of known categories. 
    (b) Utilizing state-of-the-art monocular depth and normalized object coordinates (NOCs) predictors, we obtain the depth $d_i$ and the 3D canonical point $\mathbf{p}_i$ for each observable 2D image point $\mathbf{x}_i$ on the objects.
    (c) The correspondences $(\mathbf{x}_i, d_i, \mathbf{p}_i)$ are constrained by the unknown intrinsic parameters and the object's pose.
    (d) The proposed focal length minimal solver $f$COP. For any two points on the object, the distance in the camera frame and in 3D space is determined solely by the focal length and object scale. By using a triplet of correspondences, the focal length can be estimated.}
    \label{fig:pipeline}
    \vspace{-0.5cm}
\end{figure}

The learning-based tasks of monocular depth and category-level object shape prediction have priory been combined towards the solution of 3D object pose estimation~\cite{gumeli2022roca}. 
More specifically, by combining depth predictions with category-level object priors like normalized object coordinate predictions~\cite{wang2019normalized}, the mere identification of a similarity transformation is sufficient to identify the scale and pose of the observed object. However, the focal length in such derivation, although identifiable through a relatively simple algorithm, is mostly assumed to be known in advance. Furthermore, recent works
propose learning-based methods in order to regress the focal length~\cite{zhu2024tame,piccinelli2024unidepth}, again in spite of the existence of an elegant and simple geometric solution to this problem.

In summary, our contributions are as follows:
\begin{itemize}
    \item We propose a simple and efficient minimal solver for focal length estimation, which, to the best of our knowledge, is the first to utilize category-level object priors and monocular depth estimation. 
    \item The experiments with synthetic and real-world data demonstrated the effectiveness and robustness of the proposed method. Moreover, it outperforms the current state-of-the-art monocular focal estimation methods.
\end{itemize}

\section{Related Works}
\label{sec:relatedWorks}
\noindent\textbf{Monocular Depth Estimation}
A monocular depth estimation model with good accuracy and generalization has always been the pursuit goal in this field and is also an important requirement of our problem. The early works~\cite{eigen2014depth,bhat2021adabins} in this field were all on datasets with ground truth depth labels, directly using the labels to supervise the predicted depth for training the model. However, limited by the size of the training set and the domain gap between different training sets, the generalization of this type of work is very poor. Some recent works, such as ~\cite{ranftl2020towards}, predict relative depth and use the affine-invariant loss to overcome the scale differences of different training data, thus enabling training on a larger-scale dataset and improving the generalization of the model. ~\cite{bhat2023zoedepth,yin2023metric3d} successfully transferred relative depth to metric depth. The later ~\cite{yang2024depth} effectively utilizes unlabeled images for training, further expanding the scale of the training set to enhance the zero-shot performance of the model. There are also some works~\cite{ke2024repurposing,fu2024geowizard} based on the pre-trained diffusion model to generate depth maps. The 3D world priors obtained from the large-scale pre-training of the diffusion model give it good performance in zero-shot depth estimation.

\noindent\textbf{Learning category-level object priors}
Category-level object priors, also known as canonical object representation, is the key concept of category-level object pose estimation (COPE) problem. The earliest work, ~\cite{wang2019normalized}, directly regresses the normalized object coordinates (NOCs) as object priors from RGB images, which is used in the geometric registration~\cite{umeyama1991least} with the back-projected object points to estimate the object pose. Later works, such as ~\cite{ikeda2024diffusionnocs}, utilize the pre-trained diffusion prior and normal maps to improve the quality of NOCs. Since the mainstream works~\cite{lin2023vinet,liu2023istnet,chen2024secondpose,di2022gpv,chen2021fs} are generally carried out on calibrated RGB-D images, object points are also a very common input for them. They fuse the geometric information of the object points with the semantic information in the RGB image to obtain better canonical representations.

\noindent\textbf{Monocular Focal Estimation}
The traditional monocular focal estimation methods~\cite{zhang2000flexible,zhang2004camera,wildenauer2012robust} require strong assumptions, such as those related to scene geometry or specific objects, and are not applicable to the COPE task. With the development of learning-based methods~\cite{workman2015deepfocal,hold2018perceptual,lee2021ctrl,jin2023perspective}, the demand for strong assumptions has gradually decreased, but problems of accuracy and generalization commonly exist. 
The recent work ~\cite{zhu2024tame}, which learnt the incidence field and then estimated focal length based on the plane normal constraint, has achieved good accuracy and generalization. Some other works train the focal estimation and other tasks simultaneously. For example, ~\cite{piccinelli2024unidepth} estimates the depth and camera parameters using two separate modules respectively. ~\cite{grabner2019gp2c,cifka2023focalpose++} are somewhat similar to our task. ~\cite{grabner2019gp2c} directly predicts the 2D-3D correspondences of the object and uses ~\cite{nakano2016versatile} to estimate the object pose and focal length. ~\cite{ponimatkin2022focal,cifka2023focalpose++}, knowing the CAD model of the object, utilizes a render and compare strategy to optimize the focal and object pose. However, the 2D-3D correspondence of the former is difficult to estimate precisely, and the final accuracy of the focal and pose is often not high. The operation efficiency of the latter during training and inference is very low.

\section{Method}
\label{sec:method}

\noindent\textbf{Preliminaries.}
Category-level object prior is the natural and effective information in daily life, knowledge of one category can often be seamlessly generalized to unseen intra-class objects.
The discussion for category-level objects is always entangled with NOCs, where each object is modelled in a unit cube as shown in~\cref{fig:pipeline}. It was first proposed in~\cite{wang2019normalized}, with the 2D image points of an object in the category as input, it learnt a NOCs estimator to predict the 3D point for each pixel.

\noindent\textbf{Overview.}
We propose to estimate the focal length of a single image with the category-level object priors.
\Cref{fig:pipeline} illustrates the pipeline. 
With only an image as input, we first obtain the depth and NOCs of an object by monocular depth predictor and NOCs predictor, then using the geometric relation between the 2D image and 3D NOCs, the focal length can be estimated by the proposed focal length solver $f$COP.
In this section, we formally introduce $f$COP.

\subsection{Problem Formulation}
Consider the registration between the NOCs and image points of an object, denote the given image point, the corresponding depth and NOCs as $(\mathbf{x}_i, d_i, \mathbf{p}_i)$ with $\mathbf{x}_i\in \mathbb{R}^2, d_i \in \mathbb{R}^+$ and $\mathbf{p}_i\in \mathbb{R}^3$.
Using $\tilde{\mathbf{x}}_i$ denote the homogenized coordinates of $\mathbf{x}_i$.
Then the perspective camera provides the following relation
\begin{equation}
    d_i  \mathbf{K}^{-1}\tilde{\mathbf{x}}_i + \bm{\varepsilon}_{d_i}= s \mathbf{R} (\mathbf{p}_i + \bm{\varepsilon}_{\,\mathbf{p}_i}) + \mathbf{t} + \mathbf{o}_i, \label{eq:object-pose}
\end{equation}
where $\mathbf{K}$ is the unknown camera intrinsic matrix, $s, \mathbf{R}, \mathbf{t}$ are the unknown scale, rotation and translation, $\bm{\varepsilon}_{d_i}$ and $\bm{\varepsilon}_{p_i}$ models the image depth noise and NOCs noise respectively, and $\mathbf{o}_i$ is a vector of zeros if the correspondence is an inlier or an arbitrary vectors for outlier correspondences. 

Furthermore, we assume the noise is unknown but bounded. The distribution of the noise is not inherently random due to the biased inference by the neural networks. However, we could still analyze the performance of our algorithm with the unknown but bounded noise. Formally, we assume the NOCs noise and depth noise are norm-bounded
\begin{equation}
     \|\bm{\varepsilon}_{d_i}\| \leq \delta_{d_i}, \; \|\bm{\varepsilon}_{\,\mathbf{p}_i}\| \leq \delta_{\,\mathbf{p}_i}.
\end{equation}
We will show that with the bounded noise, the estimated focal length is also bounded.
In this paper, we assume the pinhole camera model with the optical centre at the image centre.

\subsection{Focal Length and Scale Decoupling}
We propose a general approach to decouple the estimation of focal length and object scale from object pose estimation. The insight is the geometry invariant property helps to reduce the unknowns. For the simplicity of introducing the method, we only consider inlier pairs in this subsection. And we start with the noise-free case.

\noindent\textbf{Translation Elimination.}
First, we eliminate the translation in \cref{eq:object-pose} based on a common observation: the relative position between two points is invariant to object translation. To be specific, given any $2$ correspondences $(\mathbf{x}_i, d_i, \mathbf{p}_i)$ and $(\mathbf{x}_j, d_i, \mathbf{p}_j)$, we can derive the following relation
\begin{equation}
    \mathbf{K}^{-1}(d_i \tilde{\mathbf{x}}_i - d_j \tilde{\mathbf{x}}_j) = s \mathbf{R} (\mathbf{p}_i  - \mathbf{p}_j ).\label{eq:trans-invariant}
\end{equation}

\noindent\textbf{Rotation Elimination.}
Second, we eliminate the rotation by taking the norm of both sides of \cref{eq:trans-invariant}. Then it becomes
\begin{equation}
    \|\mathbf{K}^{-1}(d_i \tilde{\mathbf{x}}_i - d_j\tilde{\mathbf{x}}_j) \| = s \| \mathbf{p}_i- \mathbf{p}_j  \|.\label{eq:trans-rot-invariant}
\end{equation}
For a better introduction, we formulate the camera intrinsic matrix as $\mathbf{K}=diag(f, f, 1)$. Then expand \cref{eq:trans-rot-invariant} we have
\begin{align}
    s^2 \|\mathbf{p}_i - \mathbf{p}_j \|^2 = (d_i - d_j)^2 + \|d_i \mathbf{x}_i - d_j \mathbf{x}_j\|^2/f^2. \label{eq:constraint}
\end{align}
Therefore, we can solve for the unknown $s$ and $f$ with at least $2$ equations which need $3$ correspondences. 

Given any triplet of correspondences $(\mathbf{x}_*, d_*, \mathbf{p}_*)$ indexed by $i,j,k$, the linear system is formulated as follows
{\small
\begin{align}
    \begin{bmatrix}
        \|\mathbf{p}_i - \mathbf{p}_j \|^2 & - \|d_i \mathbf{x}_i - d_j \mathbf{x}_j\|^2\\
        \|\mathbf{p}_j - \mathbf{p}_k \|^2 & - \|d_j \mathbf{x}_j - d_k \mathbf{x}_k\|^2\\
        \|\mathbf{p}_j - \mathbf{p}_k \|^2 & - \|d_j \mathbf{x}_j - d_k \mathbf{x}_k\|^2
    \end{bmatrix}
    \begin{bmatrix}
        s^2 \\ 1/f^2
    \end{bmatrix} = 
    \begin{bmatrix}
        (d_i - d_j)^2 \\ (d_j - d_k)^2 \\ (d_i - d_k)^2
    \end{bmatrix}.\label{eq:ls}
\end{align}}
We name the proposed minimal focal length solver based on~\cref{eq:constraint} as $f$COP.
Once the focal length is estimated, object poses can be found by Umeyama~\cite{umeyama1991least} (with RANSAC in real scenario) or together with predicted metric depth served as inputs for RGB-D COPE (category-level object pose estimation) methods~\cite{lin2024instance,chen2024secondpose,li2023generative}.
\begin{remark}
    To uniquely solve for the focal length from \cref{eq:ls}, the triplet of correspondences should have distinct depths and any two pairs of the image points should not be colinear with the origin, that is $\|d_i \mathbf{x}_i - d_j \mathbf{x}_j\| \neq 0$.
\end{remark}

\subsection{Robust Focal Length Estimation}
Now let us get back to the unknown but bounded noise. 
Reconsider \cref{eq:trans-rot-invariant} with presence of noise, further denote $\mathbf{p}_{ij}:=\mathbf{p}_i - \mathbf{p}_j, \bm{\varepsilon}_{p_{ij}}:= \bm{\varepsilon}_{p_i} - \bm{\varepsilon}_{p_j}, \bm{\varepsilon}_{d_{ij}}:=\bm{\varepsilon}_{d_i} - \bm{\varepsilon}_{d_j}$ then we have
\begin{equation}
    \|\mathbf{K}^{-1}(d_i \tilde{\mathbf{x}}_i - d_j\tilde{\mathbf{x}}_j) + \bm{\varepsilon}_{d_{ij}}\|  = s \| \mathbf{p}_{ij} + \bm{\varepsilon}_{p_{ij}} \|.\label{eq:trans-rot-invariant-noisy1}
\end{equation}
First, it is easy to see that the paired noises are still bounded $\|\bm{\varepsilon}_{\mathbf{p}_{ij}}\|\leq \delta_{\,\mathbf{p}_{i}}+\delta_{\,\mathbf{p}_{j}}=: \delta_{\,\mathbf{p}_{ij}}$ and $\|\bm{\varepsilon}_{d_{ij}}\|\leq \delta_{d_{i}}+\delta_{d_{j}}=: \delta_{d_{ij}}$. 
Then using the triangle inequality
\begin{equation}
    \|\mathbf{a}\| - \| \delta \mathbf{a} \| \leq \|\mathbf{a} + \delta \mathbf{a} \| \leq \|\mathbf{a}\| + \| \delta \mathbf{a} \|,
\end{equation}
we can view \cref{eq:trans-rot-invariant-noisy1} in the following form
\begin{equation}
    \|\mathbf{K}^{-1}(d_i \tilde{\mathbf{x}}_i - d_j\tilde{\mathbf{x}}_j) \| + \tilde{\varepsilon}_{d_{ij}} = s (\| \mathbf{p}_{ij}  \| +\tilde{\varepsilon}_{\,\mathbf{p}_{ij}}),\label{eq:trans-rot-invariant-noisy2}
\end{equation}
where $|\tilde{\varepsilon}_{d_{ij}}| \leq \delta_{d_{ij}}$ and $|\tilde{\varepsilon}_{\,\mathbf{p}_{ij}}| \leq \delta_{\,\mathbf{p}_{ij}}$.
And expand \cref{eq:trans-rot-invariant-noisy2} we can formulate the noisy version of \cref{eq:constraint} as follows
\begin{equation}
    s^2 (\|\mathbf{p}_{ij} \|^2 + \mathring{\varepsilon}_{\,\mathbf{p}_{ij}}) = (d_{ij}^2 + \mathring{\varepsilon}_{d_{ij}} ) + \|d_i \mathbf{x}_i - d_j \mathbf{x}_j\|^2/f^2 , \label{eq:constraint-noisy}
\end{equation}
where $\mathring{\varepsilon}_{\,\mathbf{p}_{ij}}:=2\tilde{\varepsilon}_{\,\mathbf{p}_{ij}} \|\mathbf{p}_{ij}\| + \tilde{\varepsilon}_{\,\mathbf{p}_{ij}}^2$ and $\mathring{\varepsilon}_{d_{ij}}:=2\tilde{\varepsilon}_{d_{ij}} \|\mathbf{K}^{-1}(d_i \tilde{\mathbf{x}}_i - d_j\tilde{\mathbf{x}}_j) \| + \tilde{\varepsilon}_{d_{ij}}^2$.
Note that the object is inherently bounded by its size, and consequently, $\mathring{\varepsilon}_{\,\mathbf{p}_{ij}}$ and $\mathring{\varepsilon}_{d_{ij}}$ are also bounded. Solving a linear system perturbed by bounded noise will therefore result in a bounded estimation.

With the presence of outliers, we further consider the robust focal length estimation problem as follows
\begin{equation}
\begin{aligned}
    \max_f \; &\; \sum_k \id(|f_k - f| < \epsilon), 
\end{aligned}
\end{equation}
where $f_k$ is the estimated focal length from linear system~\cref{eq:ls} formed with each random triplet and $\epsilon$ is a noise bound. 
To solve such a 1D consensus maximization (CM) problem, voting-based methods are a popular choice due to their efficiency and optimality~\cite{bujnak2009robust,bustos2017guaranteed,yang2020teaser,peng2022arcs,zhang2024accelerating}. We adopt Interval Stabbing~\cite{peng2022arcs,zhang2024accelerating} to find the best estimated focal length for its global optimality and $\mathcal{O}(n \log n)$ complexity.

\subsection{Frame-wise focal length consistency}
\begin{figure}
    \centering
    \begin{subfigure}[t]{0.45\linewidth}
        \centering
        \includegraphics[width=\linewidth]{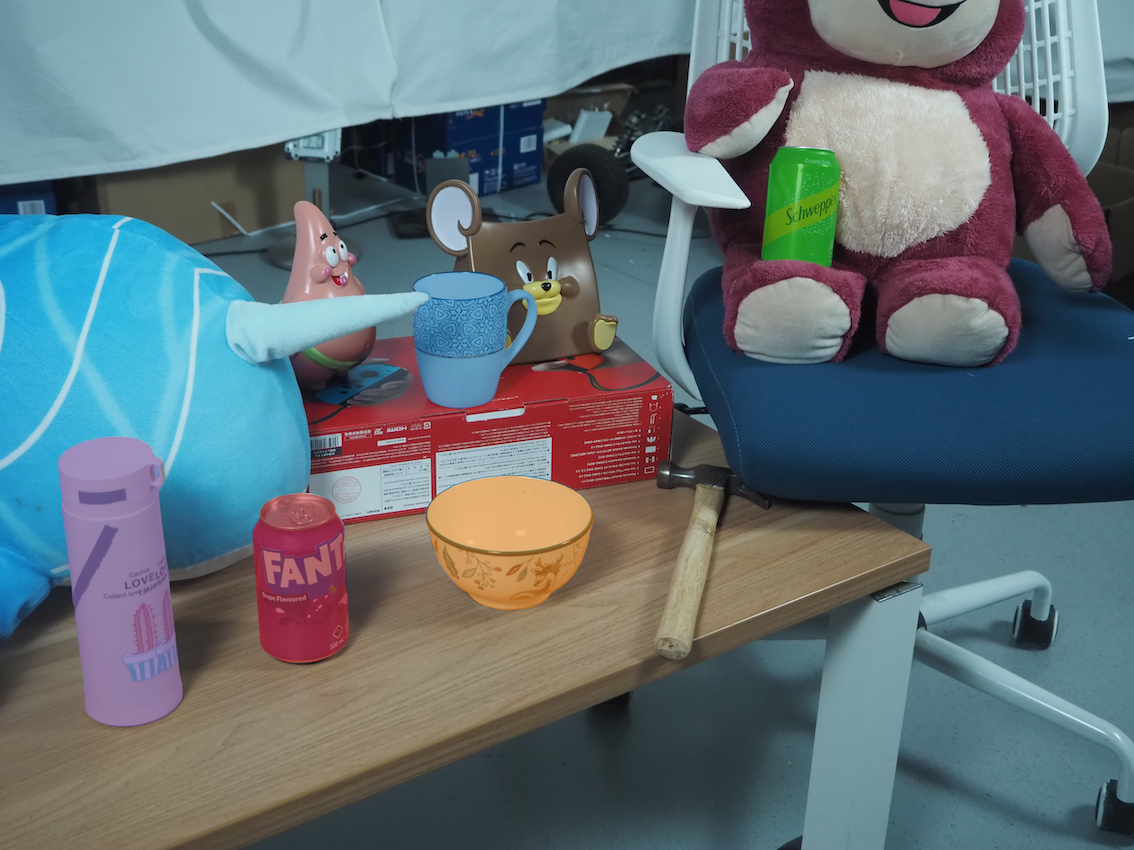}
        \caption{ {\small Objects in one frame}}
        \label{fig:method-mask}
    \end{subfigure}
    \begin{subfigure}[t]{0.45\linewidth}
        \centering
        \includegraphics[width=\linewidth]{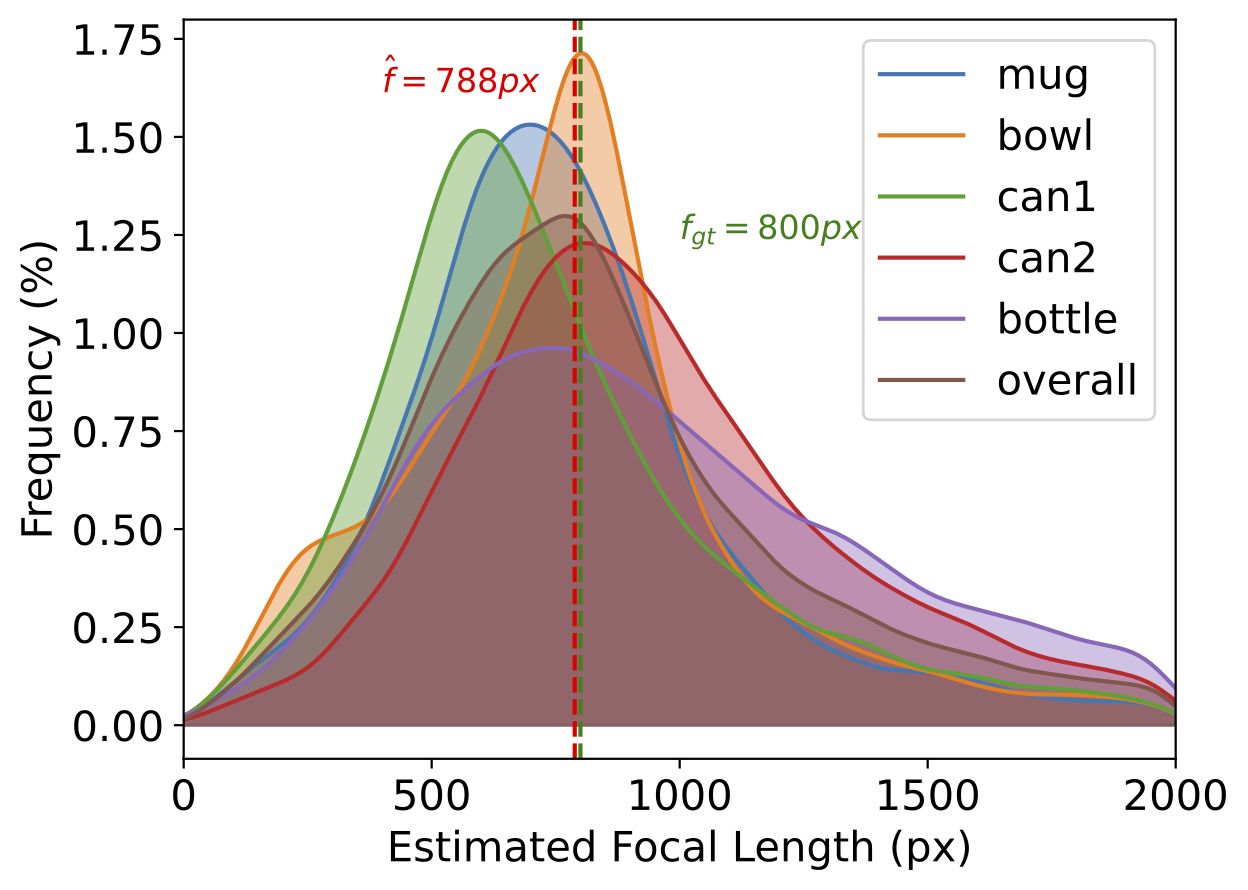}
        \caption{{\small Focal distribution}}
        \label{fig:method-density1}
    \end{subfigure}
    \caption{Scene-wise focal estimation illustration. When there are multiple objects scattering in one image frame, since the depth and NOCs could contaminated with noise, focal length estimated for each object would centered on a perturbed focal and resulting in different focal lengths from different objects. Using Interval Stabbing over focal lengths estimated from all the objects would provide a consistent and accurate estimation.}
    \label{fig:IIIC}
    \vspace{-0.2cm}
\end{figure}

As demonstrated in~\cref{fig:IIIC}, there could be multiple objects presented in one image frame, and individually estimating focal lengths for each object would result in inconsistency. Therefore, we propose to collect sets of estimated focal lengths for each object and then jointly estimate one common focal length using Interval Stabbing.

\Cref{alg:alg} presents the steps to robustly estimate focal length over a frame using multiple objects.
\begin{algorithm}
\caption{$f$COP-IS}\label{alg:alg}
\begin{algorithmic}[1]
\STATE \textbf{Input:} A image $I$ of $n$ objects, number of sampling $T$
\STATE \textbf{Output:} Focal Length of the image
\STATE $FocalSet \gets [ ]$
\FOR{$i = 1$ to $n$}
    \FOR{$t = 1$ to $T$}
        \STATE sample a triplet on object $i$
        \STATE $f_k \gets$ solve~\cref{{eq:ls}}
        \STATE $FocalSet$ append $f_k$
    \ENDFOR
\ENDFOR
\STATE $\hat{f} \gets$ Interval Stabbing on $FocalSet$
\STATE \textbf{return} $\hat{f}$
\end{algorithmic}
\end{algorithm}

\begin{figure}[h]
    \centering
    \includegraphics[width=0.99\linewidth]{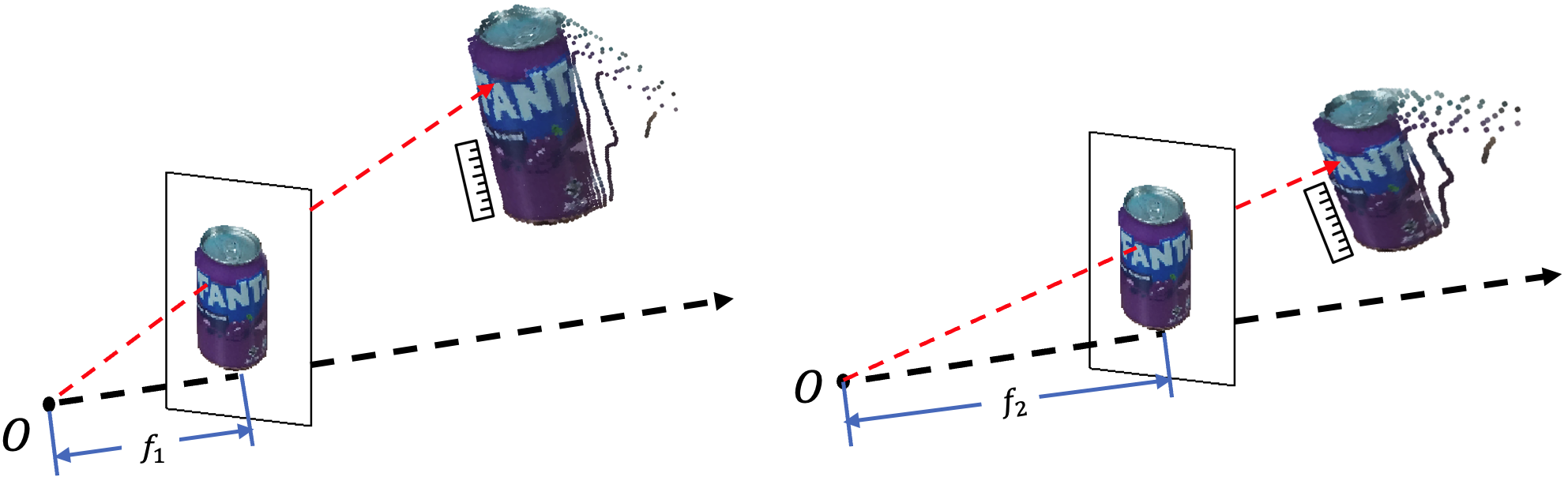}
    \caption{Same image that captured one object using different focal lengths results in different shapes of back-projected objects in 3D space. It can be observed that the object got flattened in the $x,y$-axis by using a larger focal length. Consequently, more outliers would be counted as inliers in the right one.}
    \label{fig:diff-focals}
\end{figure}

\begin{remark}
    One may ask why we do not choose to run the minimal focal length solver within Umeyama-RANSAC. The key issue arises from the fact that varying focal lengths result in different measurement scales of 3D points. Consequently, the threshold units become inconsistent for different estimated focal lengths as illustrated in~\cref{fig:diff-focals}. In the experiments section, we will compare RANSAC and the 1D CM approach to further support this observation.
\end{remark}

\section{Experiments}
\label{sec:experiments}
In this section, we evaluate the proposed focal solver $f$COP on synthetic and real-world data. 
On synthetic data, we mainly test the numerical stability of the solver, as well as the influence of depth noise and NOCs noise on the estimated focal. On real datasets, we verify that the solver can obtain accurate focal length estimation by using off-the-shelf monocular depth estimation and NOCs prediction. Since all the images in the public dataset REAL275~\cite{wang2019normalized} we tested have the same focal lengths, we additionally collected a dataset containing various different focal lengths to further verify the effectiveness and robustness of the proposed solver.

\subsection{Simulation}
We first evaluate the proposed minimal focal length solver on synthetic data.

\noindent \textbf{Setup.}
We randomly generate NOCs within a cube $[-1,1]^3$ and uniformly sample the focal length from the range $[300, 1500]$. To obtain the corresponding image points, we first apply an arbitrary rotation matrix to the NOCs and scale them within the range $[0.2, 1.0]$. Then, the NOCs are translated randomly inside a ball with a radius of $2$ meters, while the camera is positioned on the z-axis at a distance of 4 meters from the origin. 

\noindent\textbf{Metric.} We measure the focal length error and the object pose error, including translation, scale error in percentage, and rotation error in degrees. To be specific,
\begin{align}
    \mathcal{E}_{f} &\; = |\hat{f}-f_{gt}|/f_{gt} \times 100\\
    \mathcal{E}_{s} &\; = |\hat{s}-s_{gt}|/s_{gt} \times 100\\
    \mathcal{E}_{\mathbf{t}} &\; = \|\hat{\mathbf{t}}-\mathbf{t}_{gt}\|/ \|\mathbf{t}_{gt}\| \times 100\\
    \mathcal{E}_{\mathbf{R}} &\; = \arccos((\text{Tr}(\hat{\mathbf{R}}^T \mathbf{R}_{gt}) - 1 )/2).
\end{align}

\begin{figure}
    \centering
    \begin{subfigure}[t]{0.45\linewidth}
        \centering
        \includegraphics[width=\linewidth]{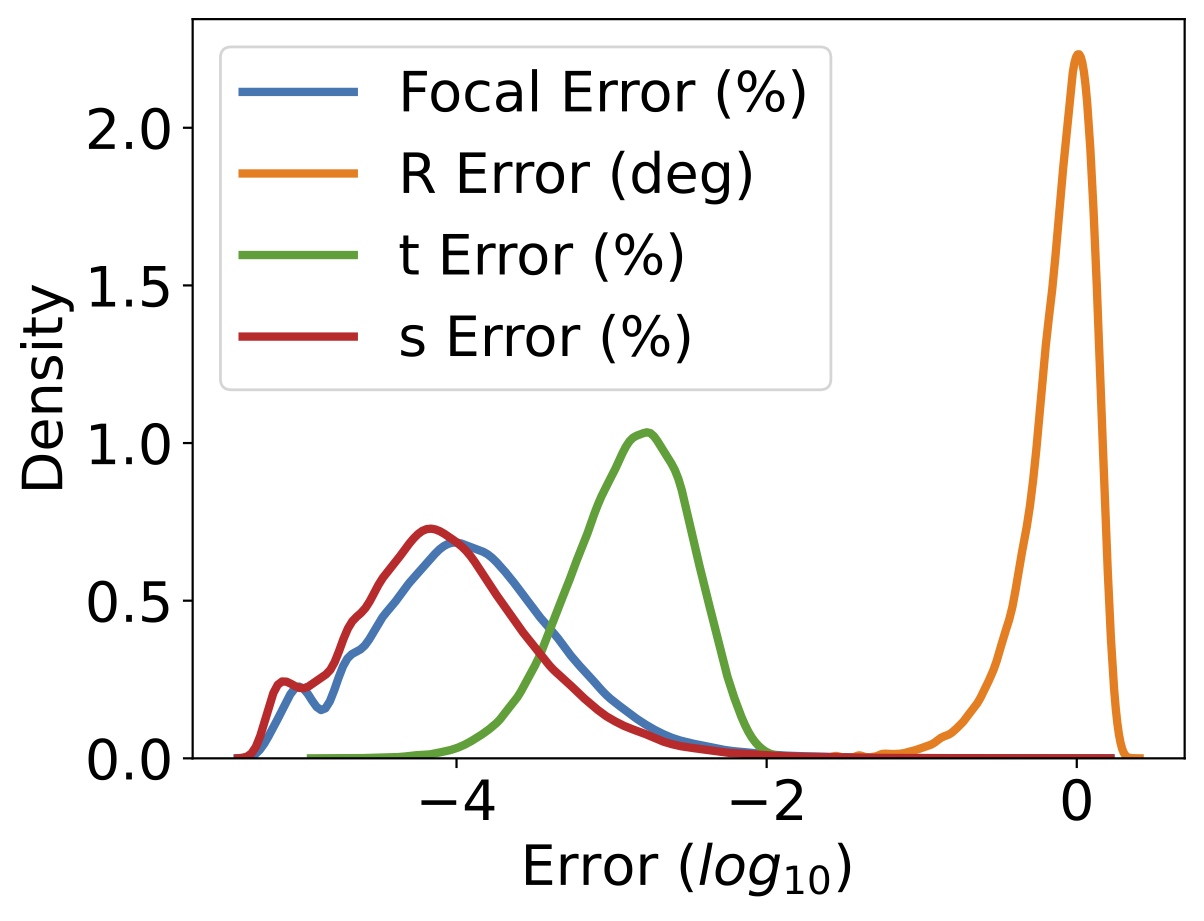}
        \caption{Numerical Stability}
        \label{fig:stability}
    \end{subfigure}
    \begin{subfigure}[t]{0.45\linewidth}
        \centering
        \includegraphics[width=\linewidth]{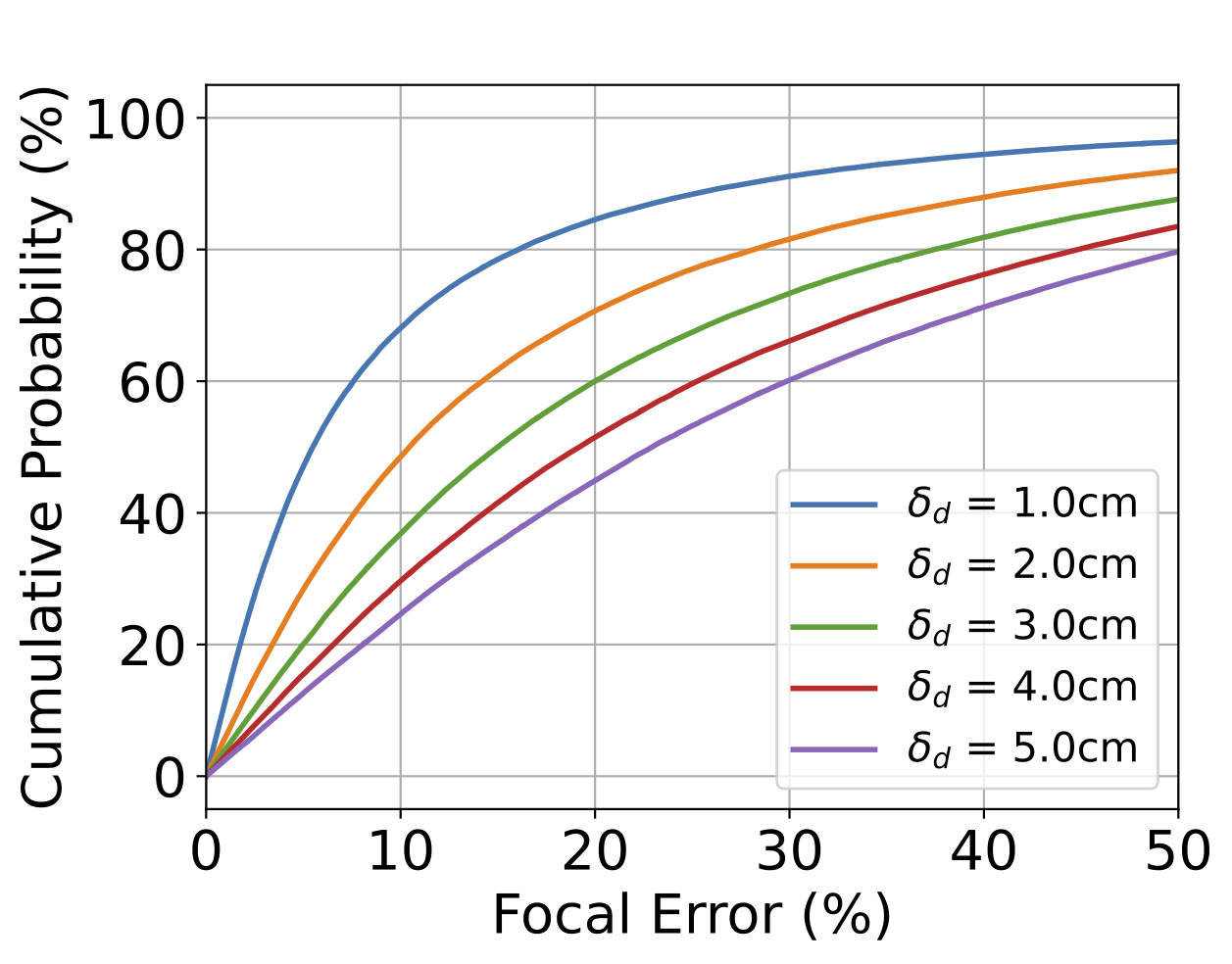}
        \caption{Depth noise}
        \label{fig:depth-noise}
    \end{subfigure}
    \begin{subfigure}[t]{0.45\linewidth}
        \centering
        \includegraphics[width=\linewidth]{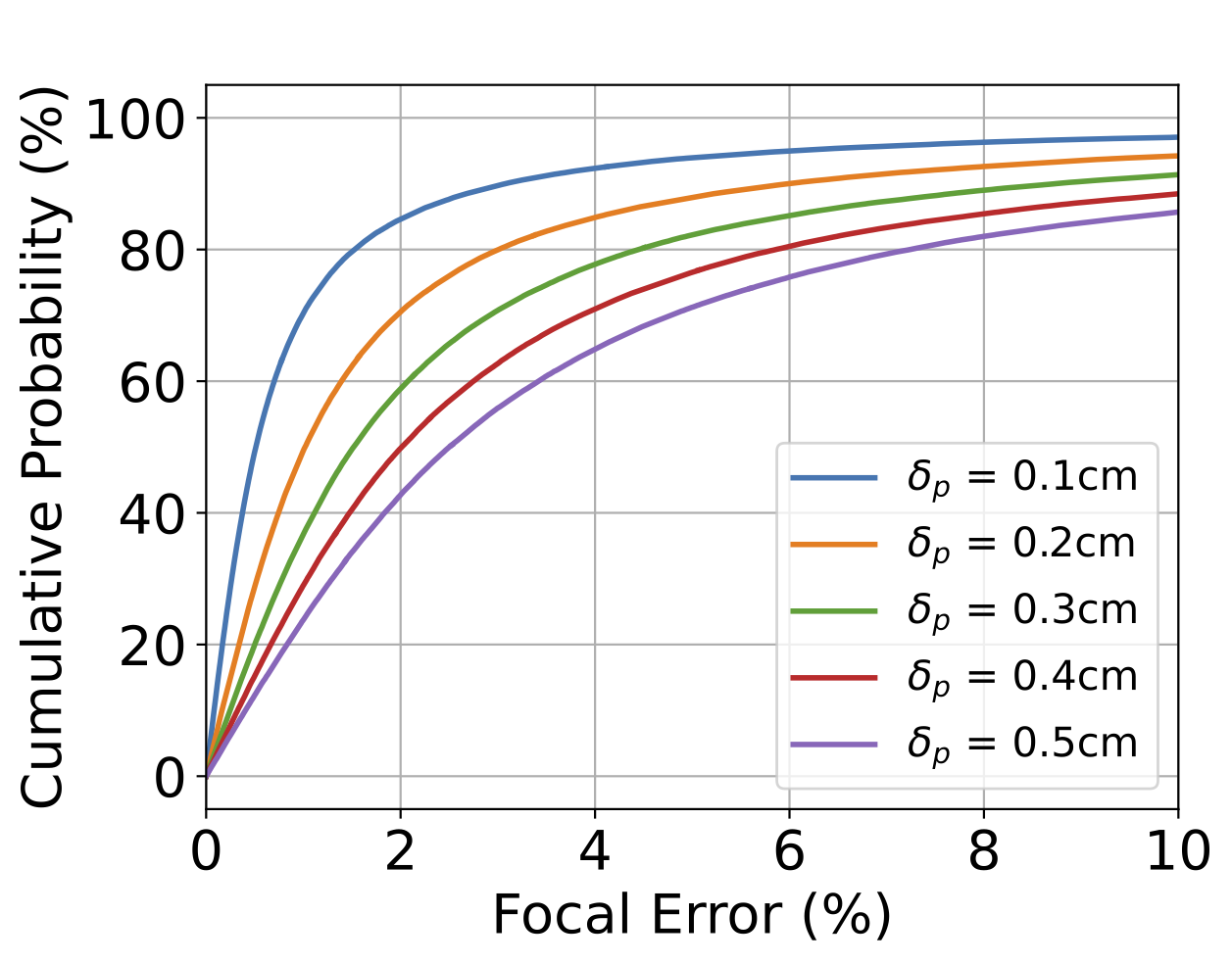}
        \caption{NOCs noise}
        \label{fig:nocs-noise}
    \end{subfigure}
    \begin{subfigure}[t]{0.45\linewidth}
        \centering
        \includegraphics[width=\linewidth]{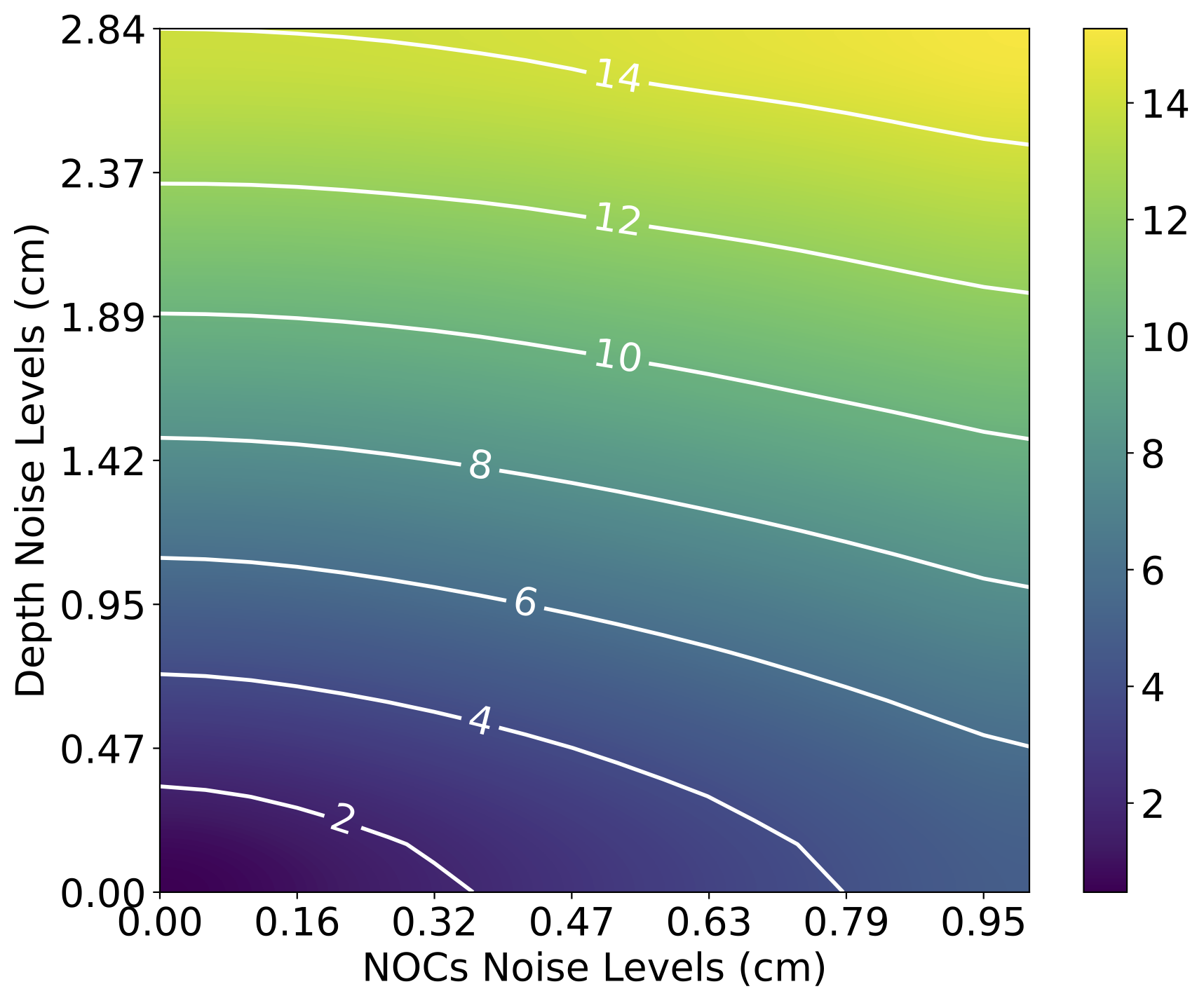}·
        \caption{Depth $\times$ NOCs noise}
        \label{fig:nocs-depth}
    \end{subfigure}
    \caption{Simulation results. The focal length solver holds great stability on noise-free data and the corresponding poses estimated by Umeyama are also stable. The focal length errors perform as the function of depth and NOCs noise.}
    \label{fig:simulation}
    \vspace{-0.5cm}
\end{figure}

\noindent \textbf{Numerical Stability.}
We randomly generate $10^5$ trials to test the stability of the proposed minimal focal length solver \cref{eq:ls} on noise-free data. \Cref{fig:stability} shows the distribution of focal length and pose errors, demonstrating that the solver provides consistently accurate estimates.

\noindent \textbf{Noise Resiliance.}
There could be two factors associated with noise as we analyze in~\ref{sec:method}, predicted depth and NOCs. To investigate the impact of noise, we randomly generate bounded depth noise and bounded NOCs noise. \Cref{fig:depth-noise,fig:nocs-noise,fig:nocs-depth} presents the performance of the proposed focal length solver under increasing noise levels. 


\subsection{Real Experiments}
We test the proposed robust focal solver $f$COP-IS and $f$COP-RANSAC on two real-world datasets and make comparisons with other state-of-the-art methods.

\begin{figure}
    \centering
    \includegraphics[width=0.7\linewidth]{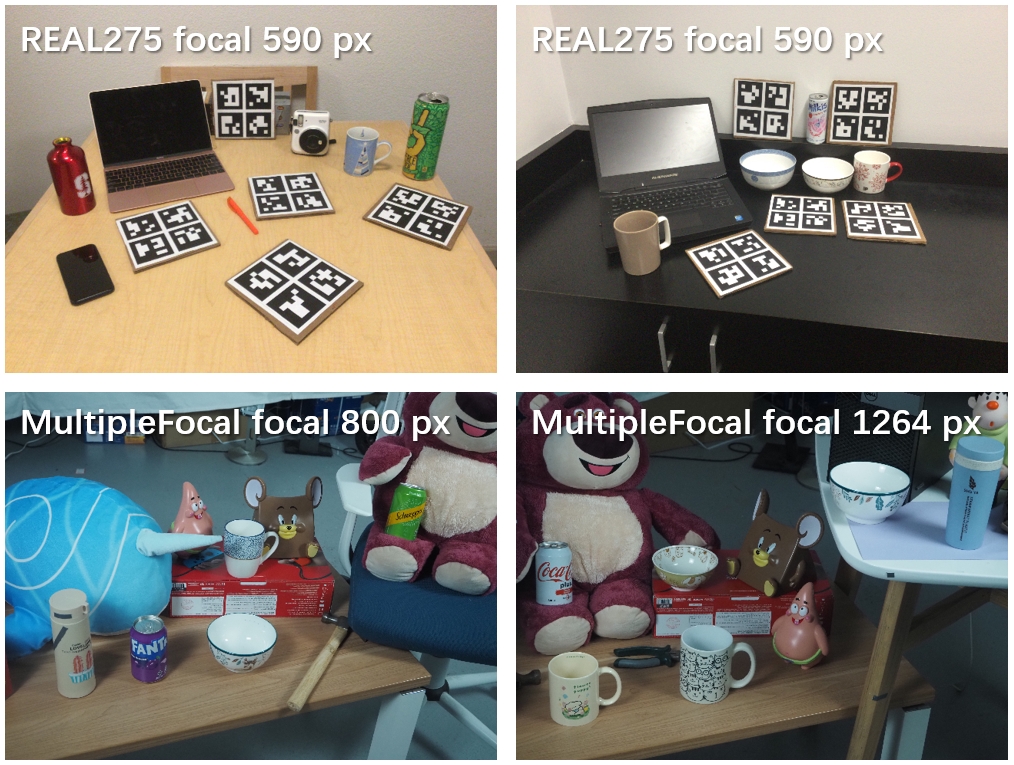}
    \caption{Example images from real datasets. The first row is images from REAL275~\cite{wang2019normalized}, there are large planes in the scene all using the same focal length $590px$. The second row is images from the proposed dataset -- MultiFocals. As its name suggested, the dataset is collected with $31$ different focal lengths ranging from $533$px to $1390$px in daily scenes.}
    \label{fig:real-imgs}
    \vspace{-0.5cm}
\end{figure}

\noindent\textbf{Dataset}-\textit{REAL275.}
It is the validation set for the NOCs predictor. Also, its scenes are also relatively close to the training sets of the SOTA monocular depth estimation methods.
There are $6$ indoor scenes with $6$ object categories: can, bowl, mug, bottle, laptop and camera, a total of $2854$ $480\times 640$ RGB-D image frames with ground truth object pose and NOCs labels. \Cref{fig:real-imgs} shows some sample images from REAL275.

\begin{figure}[t]
    \centering
    \begin{subfigure}[t]{0.45\linewidth}
        \centering
        \includegraphics[width=\linewidth]{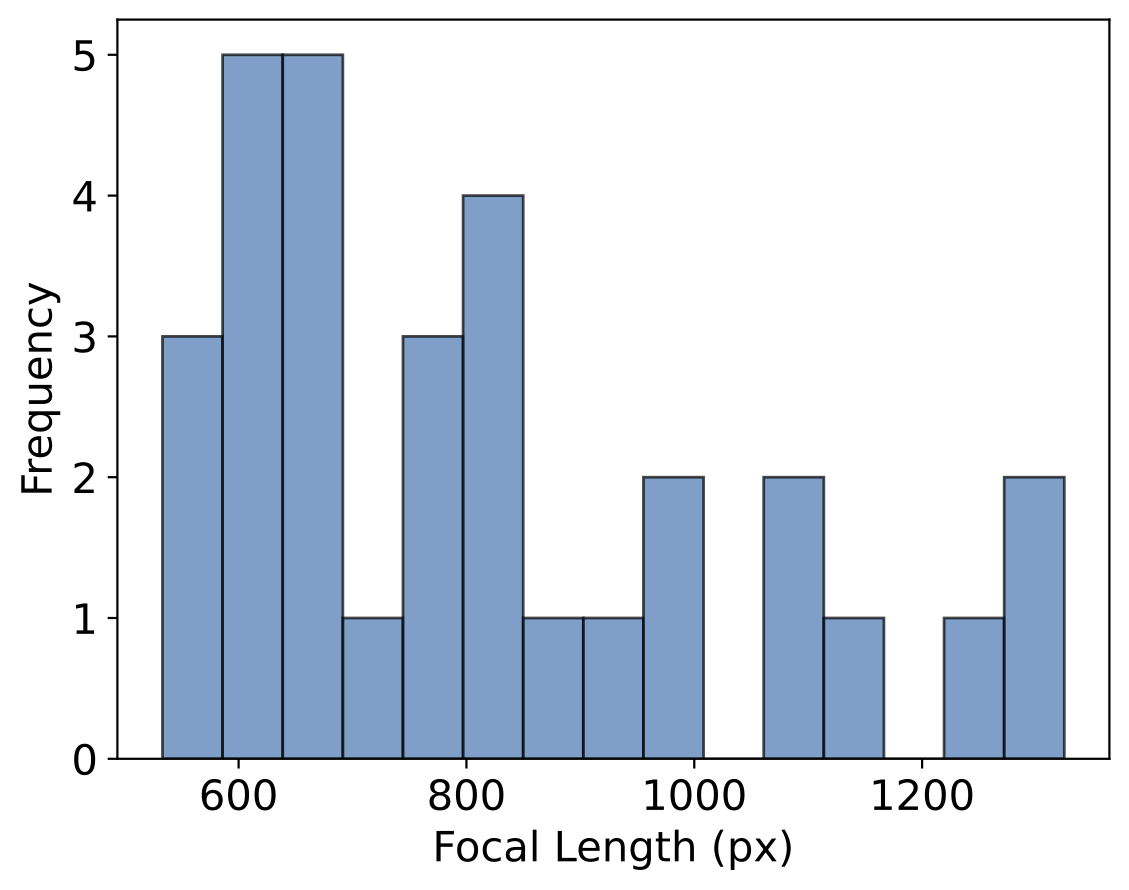}
        \caption{{\footnotesize GT Focal Length Distribution}}
        \label{fig:GT-focal}
    \end{subfigure}
    \begin{subfigure}[t]{0.45\linewidth}
        \centering
        \includegraphics[width=\linewidth]{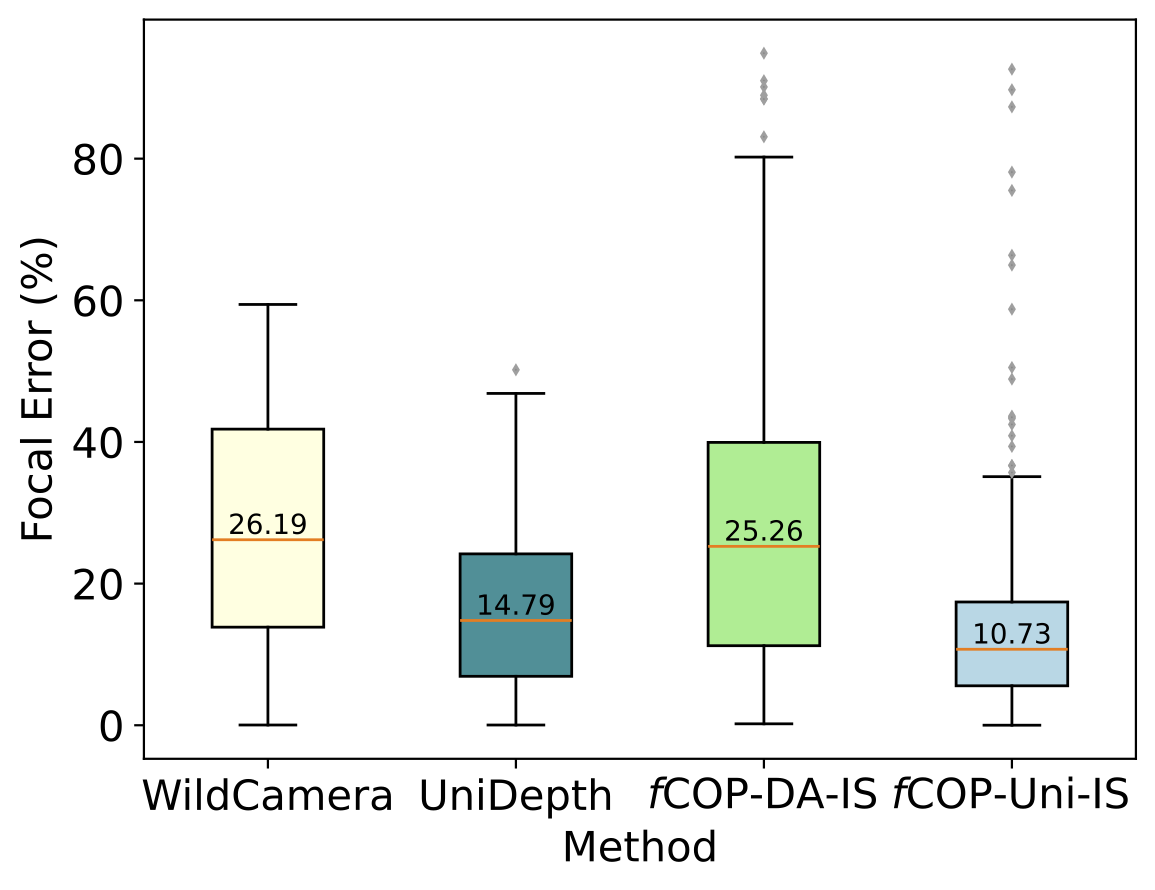}
        \caption{{\footnotesize Predicted Focal Length Errors}}
        \label{fig:boxplot-focal}
    \end{subfigure}
    \caption{Ground truth focal length distribution on MultiFocals dataset and the focal length estimation results on it. The proposed $f$COP-Uni-IS presents the best performances whereas the monocular intrinsic estimation baselines have worse performance on this out-of-domain dataset.  }
    \label{fig:real-self-collected}
    \vspace{-0.5cm}
\end{figure}

\noindent\textbf{Dataset}-\textit{MultiFocals.}
As a supplement, we additionally collected a dataset called \textit{MultiFocals}. Different from \textit{REAL275} which havs only one focal length, the images in this dataset have various focal lengths within different ranges, which can further test the robustness of the algorithm. In details, it contains $738$ $480\times 640$ image frames with $31$ different focal lengths ranging in $[533, 1390]$. Including $4$ categories(can, bowl, mug, bottle) in $10$ scenes. 
\Cref{fig:real-imgs} presents sample images from MultiFocals. \Cref{fig:GT-focal} presents the distribution of focal lengths in self-collected data.

\noindent\textbf{Setup.}
As shown in~\cref{fig:pipeline}, data preparation for the focal length estimation involves two steps: monocular depth prediction and NOCs prediction.
For monocular depth prediction, we employed two state-of-art zero-shot monocular depth methods to compare: UniDepth\cite{piccinelli2024unidepth} and Depth Anything\cite{yang2024depth}. The former estimates focal length and depth at the same time, and the latter estimates depth only.
For NOCs prediction, we follow the state-of-art DiffusionNOCS~\cite{ikeda2024diffusionnocs} pipeline using normals predicted by GeoWizard \cite{fu2024geowizard}.

To the best of our knowledge, $f$COP is the first monocular focal length solver to estimate focal length with category-level object priors, we compared the proposed $f$COP with other learning based monocular intrinsic estimation approaches.
As introduced above, \textit{UniDepth} would be a natural competitor. Additionally, we add the state-of-the-art monocular focal length estimation approach named \textit{WildCamera} \cite{zhu2024tame} for comparison. Note that both UniDepth and WildCamera use the whole scene image for intrinsic estimation, which is not a fair comparison for the proposed intrinsic solver, since we only use small fractions of image -- objects.

In the real scenarios, we wrap the proposed minimal $f$COP solver within the robust scheme. As introduced in the previous section, we propose to combine $f$COP with Interval Stabbing as described in~\cref{alg:alg} to release ourselves from the focal-threshold ambiguity. We also present results wrapping our minimal solver within the most well-known robust scheme -- RANSAC.
Given that there are several flexible choices for $f$COP on the monocular metric depth estimation and robust scheme, we name the variants in the following manner: $f$COP-\textit{depth name}-\textit{robust scheme}. The noise bound is set to be $5 px$.



\noindent\textbf{Results on REAL275.}
\Cref{tab:Median-Focal-Error} presents the comparison between the proposed $f$COP and SOTA monocular intrinsic estimation approaches. 
It can be observed that $f$COP-Uni, running $f$COP with depth predicted by UniDepth, provides the best performance on focal length estimation and outperforms the SOTAs on $5$ of out $6$ scenes. 
It is also worth noting that REAL275 can be approximately viewed as in-domain data for both WildCamera and UniDepth, given the fact that they have trained on datasets ~\cite{dai2017scannet} using the Structure Sensor~\cite{structure_sensor}. What's more, WildCamera and UniDepth also use the background information but the proposed $f$COP works only on objects and still outperforms them.

We also compare the performance of $f$COP using different monocular depth estimation approaches. 
Depth estimated by UniDepth enables $f$COP research a better estimation than Depth Anything.
Although the focal estimated by UniDepth is not as accurate as ours, its depth still served as a proper estimation when
combined with the object prior as evidenced by $f$COP-Uni.
Additionally, running $f$COP with Interval Stabbing demonstrated a clear improvement over RANSAC, as remarked in the previous section, since the threshold unit is not a problem for the rotation-translation invariant constraint~\cref{eq:constraint}.

\begin{table}[h]
    \caption{Median focal length error (\%) across all $6$ scenes in REAL275. $f$-COP-Uni-IS, the proposed $f$COP using depth predicted by UniDepth, shows the best performance and outperforms the monocular intrinsic estimation baselines.}
    \label{tab:Median-Focal-Error}
    \centering
    {\footnotesize
    \setlength{\tabcolsep}{3pt}
    \begin{tabular}{c|rrrrrrr}
        \hline
        \textbf{Method} & \textbf{Sce.1} & \textbf{Sce.2} & \textbf{Sce.3} & \textbf{Sce.4} & \textbf{Sce.5} & \textbf{Sce.6} & \textbf{All.}\\
        \hline
        WildCamera & 13.31 & \textbf{4.07} & 10.34 & 10.67 & 5.06 & 10.01 & 8.37\\
        UniDepths  & 11.17 & 9.11 & 6.93 & 9.80 & 5.52 & 11.43 & 9.09 \\
        \hline
        $f$COP-DA-RANSAC & 9.51 & 8.59 & 9.67 & 7.53 & 8.98 & \textbf{6.77} & 8.51 \\
        $f$COP-Uni-RANSAC & \textbf{6.83} & 7.27 & 7.82 & 7.73 & 8.08 & 8.22 & 7.65 \\
        $f$COP-DA-IS   & 12.31 & 8.49 & 10.29 & 11.08 & 7.42 & 7.58 & 8.54 \\
        $f$COP-Uni-IS  & 7.37 & 9.30 & \textbf{6.85} & \textbf{6.02} & \textbf{4.37} & 8.27 & \textbf{6.87} \\
        \hline
    \end{tabular}
    }
\end{table}

\noindent\textbf{Results on MultiFocals.}
Furthermore, we present results on the out-of-domain dataset, which means the WildCamera and UniDepth were not trained on focal lengths used in MultiFocals. 
\Cref{fig:boxplot-focal} shows the comparison between $f$COP-DA, $f$COP-Uni and WildCamera and UniDepth. The performances of SOTAs drop largely on the out-of-domain data. 
As expected, $f$COP-Uni presents the median focal length errors around $10.73\%$, while 
WildCamera shows a focal length error of $26.19\%$ and UniDepth shows a better error of $14.79\%$ for its geometric invariant loss helping.
As a geometric solver, the performance of $f$COP is consistently accurate.
For the WildCamera, which estimates the focal lengths based on the plane normal constraint, did not work well on MultiFocals as it did on REAL275 for $2$ reasons. 
First, it learnt the incidence field using the neural network restricts it with the domain issue. Second, it relies on the plane assumption. 
In summary, $f$COP-Uni-IS generalizes well in both in-domain and out-of-domain data.

\section{Ablation Study with REAL275}
In order to better understand the effect of predicted depth, NOCs on the focal length estimation, we conduct the following $2$ experiments using depth provided by depth sensors and ground truth NOCs of REAL275.

First, we test $f$COP on REAL275 using sensor depth. Denote the sensor depth as SD, $f$COP-SD-RANSAC and $f$COP-SD-IS show $6.58\%$ and $3.83\%$ median focal length error respectively. It further proves that $f$COP works better with IS than RANSAC. Combining the results in~\cref{tab:Median-Focal-Error}, we see that better depth would greatly lead to a better estimation of focal lengths.
Next, we replace the DiffusionNOCS prediction with ground truth NOCs provided along with REAL275. It can be observed by comparing~\cref{tab:Median-Focal-Error,tab:gt-NOCS}, that Diffusion NOCs served as a high-quality estimation for category-level objects.

\begin{table}[h]
\caption{Ablation study on REAL275 using ground truth NOCs.}
\label{tab:gt-NOCS}
\centering
\vspace{-0.5cm}
\begin{tabular}{c|c}
\hline
\textbf{Method} & $\mathcal{E}_{f} (\%)$ \\ \hline
$f$COP-SD-RANSAC & 3.80 \\
$f$COP-Uni-RANSAC & 5.96 \\ 
$f$COP-SD-IS & 3.15 \\
$f$COP-Uni-IS & 9.12 \\ \hline
\end{tabular}\vspace{-0.2cm}
\end{table}

\section{Application on Category-Level Object Pose Estimation}
\label{sec:application}
\begin{figure}
    \centering
    \includegraphics[width=0.95\linewidth]{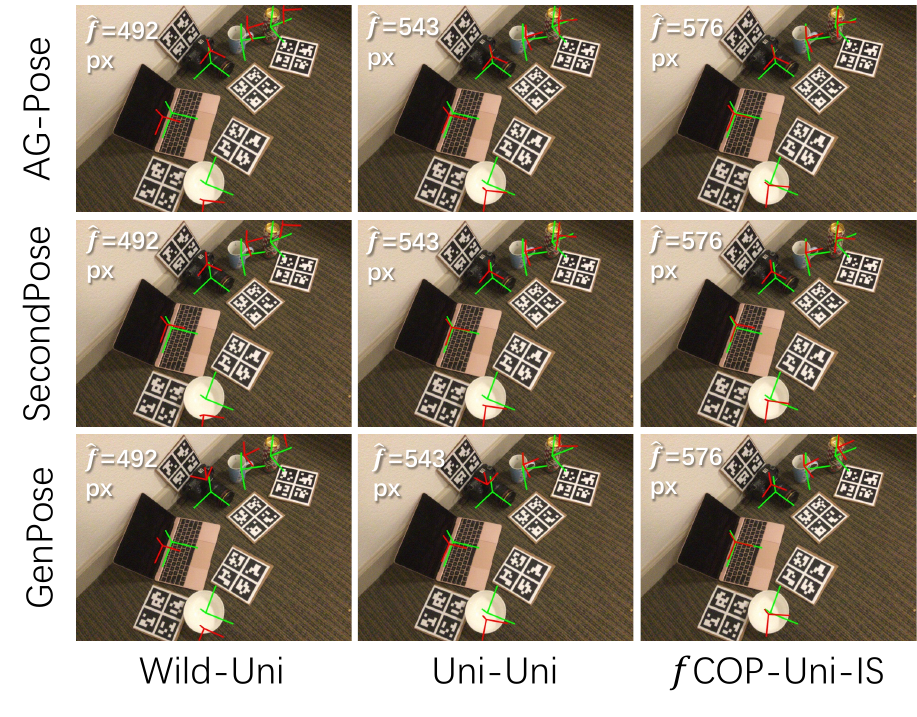}
    \caption{
    Uncalibrated object pose qualitative visualization on REAL275. Green lines on each object represent the ground truth rotation and position and red lines for estimated. 
    The ground truth focal length is $590$px and the estimated focal length is listed in the left corner of each image.
    The accurate focal length estimated by $f$COP-Uni-IS, further lead to accurate object pose estimation using both $3$ RGB-D COPE methods.}
    \label{fig:REAL275-qualitative}
    \vspace{-0.7cm}
\end{figure}

With the camera focal length estimated, object poses can be further estimated. We adopt the state-of-the-art RGB-D category-level object pose estimation methods to do the downstream test, where we use estimated metric depth and estimated focal to replace the depth provided by RGB-D as inputs for the network. We show the rotation error and translation error in~\cref{tab:ablation-RGBD-transposed}. Note that the estimated metric depth has a scale error which makes the estimated object position corrupted by it. Then we turn to present the angular translation error, which measures the angle between the estimated translation and the ground truth. We can observe that the $f$COP-Uni-IS demonstrates the best translation errors for all $3$ SOTA RGB-D COPE methods and comparable rotation errors.
We further provide the qualitative results in~\cref{fig:REAL275-qualitative}, the accurate focal length estimated by $f$COP-Uni-IS resulting in the best object translation estimation for each object.

\begin{table}[h!]
\centering
\caption{Ablation with RGB-D category level object pose network.}
\label{tab:ablation-RGBD-transposed}
{\footnotesize
    \setlength{\tabcolsep}{3pt}
    \begin{tabular}{l|rr|rr|rr}
    \hline
    \multirow{2}{*}{\textbf{Method}} & \multicolumn{2}{c|}{AG-Pose~\cite{lin2024instance}} & \multicolumn{2}{c|}{SecondPose~\cite{chen2024secondpose}} & \multicolumn{2}{c}{GenPose~\cite{li2023generative}} \\ \cline{2-7}
     & $\mathcal{E}_{\mathbf{R}}(^\circ)$ & $\mathcal{E}_{\mathbf{t}}(^\circ)$ & $\mathcal{E}_{\mathbf{R}}(^\circ)$ & $\mathcal{E}_{\mathbf{t}}(^\circ)$ & $\mathcal{E}_{\mathbf{R}}(^\circ)$ & $\mathcal{E}_{\mathbf{t}}(^\circ)$ \\ \hline
    Wild-Uni        & 5.17           & 1.24 & 5.11           & 1.26 & 5.81           & 1.44 \\ 
    Uni-Uni         & \textbf{5.04}  & 1.37 & \textbf{4.99}  & 1.38 & 5.60  & 1.59 \\ 
    $f$COP-Uni-IS   & 5.25           & \textbf{1.05} & 5.20           & \textbf{1.09} & \textbf{5.52}           & \textbf{1.30} \\ 
    \hline
    \end{tabular}
}\vspace{-0.2cm}
\end{table}

\section{Conclusion}
\label{sec:conclusion}
In this paper, we propose a novel focal length solver, which can utilize category-level object priors, along with the state-of-the-art monocular depth estimation to output focal length estimations in a closed form. Although it is limited by object categories and the current monocular depth estimation is still its performance bottleneck, the fact is that nowadays category-level object canonical representation learning is expanding towards an increasing number of categories, even in the open vocabulary. Moreover, the accuracy and generalization ability of monocular depth estimation are rapidly improving over time. It can be expected that our solver will have better accuracy and more application scenarios in the future.

{
    \small
    \bibliographystyle{IEEEtranS}
    \bibliography{IEEEexample}
}

\end{document}